\documentclass[lettersize,journal]{IEEEtran}
\usepackage{amsmath,amsfonts}
\usepackage{algorithmic}
\usepackage{algorithm}
\usepackage{array}
\usepackage[caption=false,font=normalsize,labelfont=sf,textfont=sf]{subfig}
\usepackage{mathtools}

\DeclarePairedDelimiter\floor{\lfloor}{\rfloor}

\usepackage{multirow}
\usepackage{makecell} 
\usepackage{bbding}
\usepackage{textcomp}
\usepackage{stfloats}
\usepackage{url}
\usepackage{verbatim}
\usepackage{graphicx}
\usepackage{booktabs}
\usepackage{cite}
\begin{document}

\title{GVSynergy-Det: Synergistic Gaussian-Voxel Representations for Multi-View 3D Object Detection}

\author{Yi Zhang, Yi Wang,~\IEEEmembership{Member,~IEEE}, Lei Yao, Lap-Pui Chau,~\IEEEmembership{Fellow,~IEEE}
\thanks{The research work was conducted in the JC STEM Lab of Machine Learning and Computer Vision funded by
The Hong Kong Jockey Club Charities Trust. This research received partially support from the Global STEM
Professorship Scheme from the Hong Kong Special Administrative Region.

Yi Zhang, Yi Wang, Lei Yao, Lap-Pui Chau are with
the Department of Electrical and Electronic Engineering, The Hong Kong Polytechnic University, Hong Kong, China. E-mail: yi-eee.zhang@connect.polyu.hk, yi-eie.wang@polyu.edu.hk, rayyoh.yao@connect.polyu.hk, lap-pui.chau@polyu.edu.hk.}}



\maketitle

\begin{abstract}
Image-based 3D object detection aims to identify and localize objects in 3D space using only RGB images, eliminating the need for expensive depth sensors required by point cloud-based methods. Existing image-based approaches face two critical challenges: methods achieving high accuracy typically require dense 3D supervision, while those operating without such supervision struggle to extract accurate geometry from images alone. In this paper, we present GVSynergy-Det, a novel framework that enhances 3D detection through synergistic Gaussian-Voxel representation learning. Our key insight is that continuous Gaussian and discrete voxel representations capture complementary geometric information: Gaussians excel at modeling fine-grained surface details while voxels provide structured spatial context. We introduce a dual-representation architecture that: 1) adapts generalizable Gaussian Splatting to extract complementary geometric features for detection tasks, and 2) develops a cross-representation enhancement mechanism that enriches voxel features with geometric details from Gaussian fields. Unlike previous methods that either rely on time-consuming per-scene optimization or utilize Gaussian representations solely for depth regularization, our synergistic strategy directly leverages features from both representations through learnable integration, enabling more accurate object localization. Extensive experiments demonstrate that GVSynergy-Det achieves state-of-the-art results on challenging indoor benchmarks, significantly outperforming existing methods on both ScanNetV2 and ARKitScenes datasets, all without requiring any depth or dense 3D geometry supervision (e.g., point clouds or TSDF). 
\end{abstract}


\section{Introduction}
3D object detection is a fundamental task in computer vision that aims to localize and classify objects in three-dimensional space, serving as a cornerstone for numerous applications, including autonomous driving, robotics navigation, augmented reality, and smart surveillance systems \cite{MIR-2024-09-411} \cite{10.1109/TCSVT.2024.3480691}. While 3D object detection can be performed using various input modalities, point cloud-based methods have traditionally dominated the field due to their direct access to precise geometric information. However, these approaches face significant limitations: they require expensive depth sensors (LiDAR or RGB-D cameras), suffer from limited deployment scenarios due to hardware constraints, and often struggle with sparse or incomplete point cloud data. In contrast, image-based 3D object detection has gained increasing attention as it only requires RGB cameras — ubiquitous, cost-effective sensors that can be easily integrated into various platforms. Therefore, our work focuses on advancing image-based 3D object detection to make this critical technology more accessible and practical. 

\begin{figure}[t]
\centering
\includegraphics[width=0.95\linewidth]{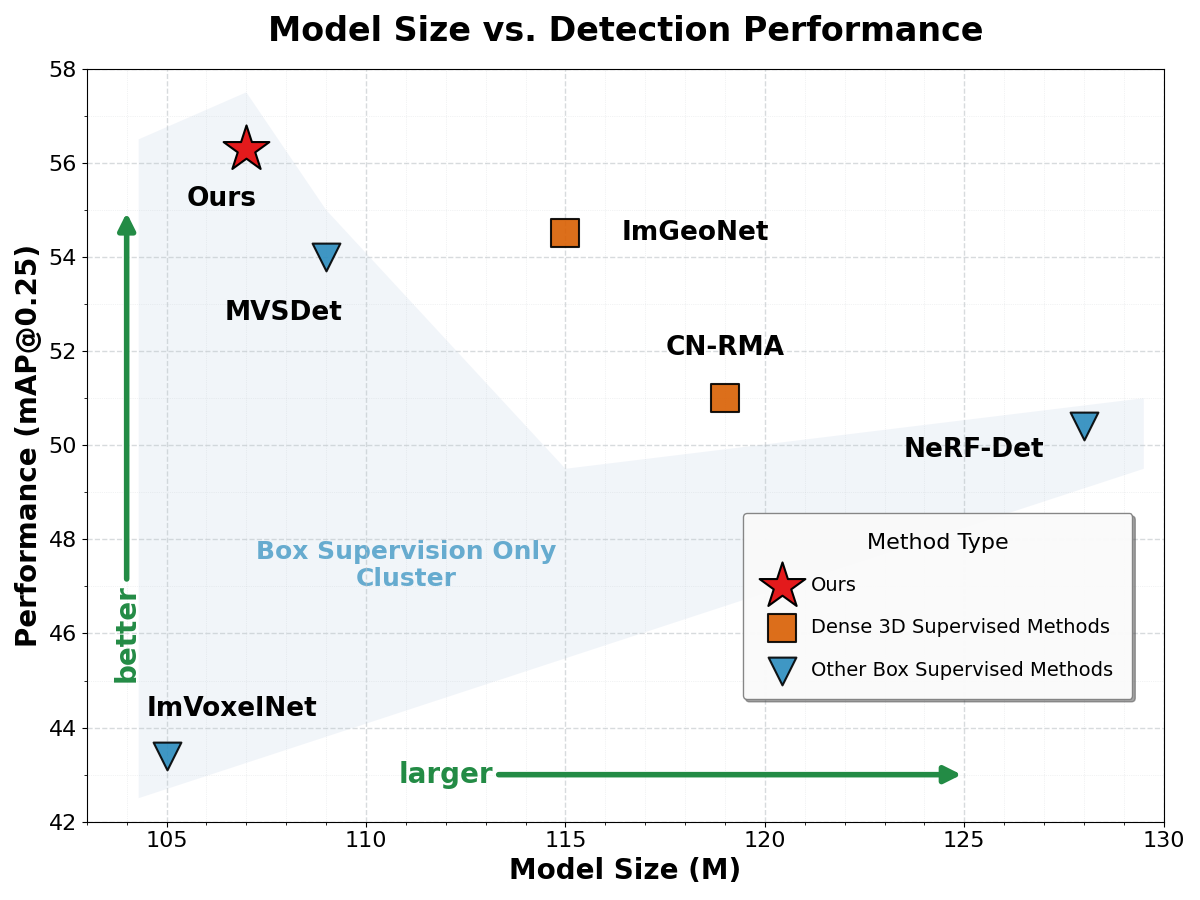} 
\caption{\textbf{Model Size vs. Detection Performance.} We compare our GVSynergy-Det (red star) against state-of-the-art methods. The x-axis represents model size (Parameters in Millions), and the y-axis represents detection performance (mAP@0.25). Our method achieves a superior Pareto frontier, offering the highest accuracy with a compact model size, notably without requiring any dense 3D supervision (e.g., point clouds or depth).}
\label{fig:model_size_vs_perf}
\end{figure}

The landscape of image-based 3D object detection methods can be broadly categorized into two paradigms based on their supervision requirements. The first category includes methods that rely on additional 3D supervision signals beyond 3D bounding box annotations. ImGeoNet \cite{tu2023imgeonet} leverages ground truth point clouds to supervise voxel occupancy prediction for improved 3D detection. CN-RMA \cite{cnmar2024} requires ground truth Truncated Signed Distance Function (TSDF) volumes to first reconstruct scenes before performing detection. More recently, 3DGeoDet \cite{11045444} utilizes depth maps to supervise the generation of intermediate 3D representations, enhancing the model's geometric understanding for detection tasks. While these methods achieve promising results, their dependence on expensive 3D annotations limits their practical deployment. 

The second category comprises methods that operate solely with 3D bounding box supervision. ImVoxelNet \cite{rukhovich2022imvoxelnet} pioneered this direction by projecting 2D features into 3D voxel space, though it struggles with depth ambiguity in the unprojection process. NeRF-Det \cite{Xu_2023_ICCV} addresses this limitation by incorporating Neural Radiance Fields (NeRF) \cite{10.1145/3503250} for geometry reasoning. However, while efficient, NeRF-Det is limited by the coarse nature of its learned geometry, often resulting in imprecise object boundaries. MVSDet \cite{10.5555/3737916.3742138} employs multi-view stereo with efficient plane sweeps to achieve strong results, while NeRF-Det++ \cite{10969554} extends NeRF-Det with enhanced sampling strategies and refined depth estimation modules to improve geometric consistency.


Despite these advances, existing methods still face fundamental limitations. For instance, while recent works have explored 3D Gaussian Splatting \cite{10.1145/3592433} for detection, 3DGS-Det \cite{cao20243dgsdetempower3dgaussian} requires time-consuming per-scene optimization, making it impractical for real-time applications. Meanwhile, methods like MVSDet \cite{10.5555/3737916.3742138} utilize the Gaussian module solely as an auxiliary depth regularizer during training, discarding the rich, fine-grained surface features encoded within the Gaussians during the actual detection process. This represents a significant missed opportunity, as the continuous Gaussian representation captures detailed surface information that is complementary to the structural context modeled by discrete voxels.

To address this limitation, we propose GVSynergy-Det, which introduces the first deep cross-representation learning framework between Gaussian and voxel representations for 3D object detection. Our key technical innovation lies in designing a synergistic integration mechanism that directly combines features from both representations. 
Unlike previous approaches that isolate these branches, we develop a cross-enhancement module that dynamically aggregates features based on their local reliability and geometric occupancy. The integration employs an adaptive weighting scheme conditioned on the fused features, allowing the network to selectively emphasize the most informative representation for each spatial location. Furthermore, by explicitly modeling occupancy from the Gaussian branch, we provide geometric guidance that effectively filters out empty space noise, ensuring that the detector focuses on regions with actual physical presence.

Extensive experiments validate the effectiveness of our approach on challenging indoor benchmarks. As illustrated in Figure \ref{fig:model_size_vs_perf}, GVSynergy-Det achieves a superior trade-off between detection accuracy and model complexity compared to existing approaches, notably achieving these results relying solely on 3D bounding box annotations, without needing depth or dense geometry supervision. On ScanNetV2 \cite{dai2017scannet}, our method achieves 56.3 mAP@0.25 and 32.1 mAP@0.50, outperforming the previous best method MVSDet by 2.3 and 3.1 points, respectively. On ARKitScenes \cite{dehghan2021arkitscenes}, we achieve 44.1 mAP@0.25 and 30.6 mAP@0.50, surpassing MVSDet by 1.2 and 3.6 points.

In summary, our contributions are threefold:
\begin{itemize}
    \item We are the first to develop a synergistic cross-representation learning framework that directly integrates generalizable 3D Gaussian Splatting with voxel-based detection at the feature level, going beyond previous works that either require per-scene optimization or use Gaussians solely for depth regularization.
    \item We propose a novel cross-representation enhancement mechanism that adaptively combines complementary geometric information from continuous Gaussian and discrete voxel representations based on local reliability and geometric occupancy, enabling more accurate object localization through learnable integration.
    \item We achieve state-of-the-art performance among box supervised methods on both ScanNetV2 and ARKitScenes benchmarks, demonstrating the effectiveness of our synergistic dual-representation approach across different indoor environments.
\end{itemize}

\section{Related Work}
\subsection{Point Cloud-based 3D Object Detection}
Point cloud-based methods have established strong foundations for 3D object detection by directly leveraging geometric information from depth or LiDAR sensors. These methods generally follow two paradigms: point-based approaches that process unstructured point sets directly, and voxel-based methods that discretize space into regular grids. PointNet \cite{qi2017pointnet} and PointNet++ \cite{10.5555/3295222.3295263} pioneered deep feature learning on unstructured point sets. Building on this, PointRCNN \cite{shi2019pointrcnn} proposed a two-stage framework generating proposals directly from points, while 3DSSD \cite{yang20203dssd} introduced a fusion sampling strategy to eliminate the computationally expensive upsampling layers, achieving efficient single-stage detection.

Voxel-based methods discretize the irregular point clouds into regular grids to apply efficient Convolutional Neural Networks (CNNs). VoxelNet \cite{8578570} introduced the concept of learning features within voxels, which SECOND \cite{yan2018second} accelerated using sparse 3D convolutions. PointPillars \cite{8954311} further improved efficiency by collapsing vertical columns into 2D pseudo-images. To balance fine-grained point features with efficient voxel processing, PV-RCNN \cite{shi2020pv} integrated multi-scale voxel features with keypoint features. More recently, fully sparse detectors like VoxelNeXt \cite{10204123} have demonstrated that detection can be performed directly on sparse voxel features without dense heads.

Transformer-based methods have recently gained attention due to their ability to capture long-range dependencies. 3DETR \cite{9711345} applies a standard transformer encoder-decoder architecture to point clouds, while VoTr \cite{9710835} introduces voxel-based local attention. DSVT \cite{10203294} further addresses the sparsity issue by employing dynamic sparse window attention, achieving state-of-the-art performance. Recent advancements in this domain have further pushed the boundaries of feature representation and proposal quality. For instance, LeadNet \cite{10124821} introduces a long-short range adaptive transformer that utilizes dynamic query sampling to adapt to varying point densities. Complementary approaches have explored semantic-aware multi-branch sampling \cite{10836835} and local-to-global semantic learning across different views \cite{10520310} to enhance context awareness, while others have successfully adapted diffusion models for voting refinement \cite{10602535}. Additionally, multi-modal approaches \cite{10109207,9826439} have emerged that fuse point cloud data with RGB images for enhanced performance. While these methods achieve impressive accuracy, they remain fundamentally limited by their dependence on expensive LiDAR or RGB-D sensors, motivating the development of image-only alternatives like our approach.

\subsection{Image-based 3D Object Detection}
Image-based approaches aim to hallucinate 3D geometry from 2D inputs, a fundamentally ill-posed problem due to depth ambiguity.

Lifting-based methods transform 2D image features into 3D representations. ImVoxelNet \cite{rukhovich2022imvoxelnet} projects 2D features into a 3D voxel volume via camera calibration matrices, treating the volume as a "virtual point cloud." However, without geometric supervision, the back-projected features are often smeared along the ray, causing depth ambiguity. To mitigate this, ImGeoNet \cite{tu2023imgeonet} utilize ground truth point clouds to supervise the occupancy of the voxel grid. CN-RMA \cite{cnmar2024} took a two-stage approach, first reconstructing 3D scenes using TSDF supervision before applying detection. 3DGeoDet \cite{11045444} converts depth predictions into intermediate 3D representations to integrate geometric information, requiring depth supervision for training. However, these methods complicate the training pipeline by requiring dense 3D geometric annotations.

NeRF-based methods leverage the differentiable rendering of Neural Radiance Fields \cite{10.1145/3503250}
to learn geometry implicitly. NeRF-Det \cite{Xu_2023_ICCV} integrates a NeRF branch to refine the feature volume by minimizing photometric error. However, NeRF-Det relies on the implicit extraction of geometry (e.g., opacity fields), which often yields noisy surfaces that lack precise boundaries for object localization. NeRF-Det++ \cite{10969554} improves upon this by introducing ordinal depth supervision and perspective-aware sampling to enforce better geometric consistency.

While recent stereo-based approaches like DSC3D \cite{10753651} have advanced 3D detection in autonomous driving, they rely on fixed-baseline constraints and outdoor priors that are inapplicable to the unconstrained camera poses and complex occlusions typical of indoor environments. Consequently, indoor approaches primarily leverage Multi-View Stereo (MVS) techniques to handle arbitrary views. 
MVSDet \cite{10.5555/3737916.3742138} constructs cost volumes via plane-sweeping and employs a probabilistic sampling mechanism to determine the placement of image features in 3D space. While MVSDet incorporates pixel-aligned 3D Gaussian Splatting, it utilizes this representation solely as an auxiliary depth regularizer during training. This approach discards the rich semantic and structural information contained within the Gaussian primitives during the actual detection process. In contrast, our GVSynergy-Det directly integrates the high-fidelity features from the Gaussian branch into the detection pipeline via a cross-representation enhancement module, further enhancing the detection performance.

\subsection{3D Gaussian Splatting and Generalizable Models}
3D Gaussian Splatting (3DGS) \cite{10.1145/3592433} represents scenes as collections of 3D Gaussians primitives, each characterized by position, covariance, opacity, and appearance attributes, enabling real-time rendering via splatting-based rasterization. While the original 3D Gaussian Splatting requires lengthy per-scene optimization, Generalizable 3D Gaussian Splatting methods have emerged to predict Gaussian parameters directly from sparse images in a feed-forward manner.

Recent approaches \cite{10655681, 10.1007/978-3-031-72664-4_21, 10656101, zheng2024gpsgaussian, 10.5555/3737916.3741325, 10965894} have successfully adapted 3DGS for instant inference across different domains. For general scene reconstruction, pixelSplat \cite{10655681} overcomes scale ambiguity and local minima by employing an epipolar transformer and predicting probabilistic depth distributions. MVSplat \cite{10.1007/978-3-031-72664-4_21} further improves efficiency by constructing a cost volume via plane sweeping to robustly localize Gaussian centers. In object-centric scenarios, Splatter Image \cite{10656101} utilizes a standard 2D U-Net to regress a "splatter image" where each pixel maps to a 3D Gaussian, achieving ultra-fast reconstruction. Distinct from general scenes, GPS-Gaussian \cite{zheng2024gpsgaussian} targets human novel view synthesis, learning human-specific priors from scan data to regress Gaussian parameter maps jointly with depth estimation. These methods demonstrate that high-quality 3D representations can be inferred instantly without per-scene optimization.

The explicit nature of Gaussian primitives facilitates their integration into downstream tasks. In semantic segmentation, methods like Gaussian Grouping \cite{Ye2023GaussianGS} and CoSSegGaussians \cite{dou2024cosseggaussians} attach semantic labels or features to 3D Gaussians, allowing 2D segmentation masks to be lifted into 3D space with high multi-view consistency. However, the potential of these Gaussian representations for 3D object detection remains under-explored. 3DGS-Det \cite{cao20243dgsdetempower3dgaussian} attempts to use Gaussians for detection but reverts to slow per-scene optimization, limiting its practicality. While MVSDet \cite{10.5555/3737916.3742138} employs generalizable Gaussians, it utilizes them solely for depth regularization, discarding the rich geometric and semantic features encoded in the Gaussian field. In contrast, our method not only employs generalizable Gaussian Splatting for supervision but also designs a cross-representation enhancement mechanism to fully exploit both geometric and semantic features from Gaussian fields, achieving superior detection performance without per-scene optimization.

\section{Methodology}
\begin{figure*}[t]
\centering
\includegraphics[width=0.999\linewidth]{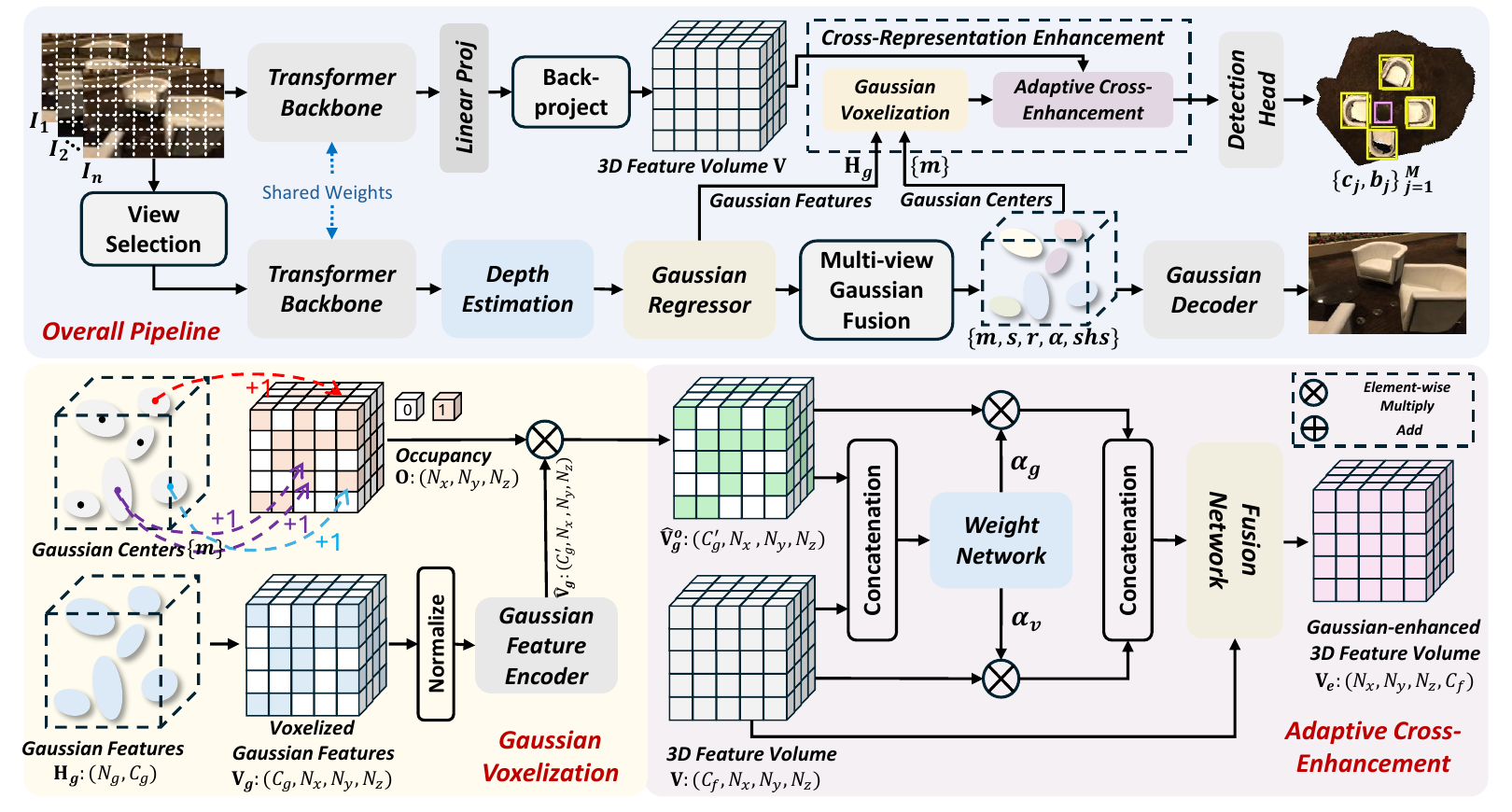} 
\caption{Overall framework of GVSynergy-Det. Given multi-view RGB images, the GVSynergy-Det first extracts features using shared transformer backbones and constructs both a 3D voxel representation through back-projection and a Gaussian representation through pixel-aligned Gaussian prediction. Then, the Cross-Representation Enhancement module is proposed to synergistically integrate the 3D voxel representation and Gaussian representation. Specifically, Gaussian Voxelization transforms irregular Gaussian features into a voxel-aligned grid with occupancy guidance, followed by Adaptive Cross-Enhancement that learns to dynamically weight contributions from both representations based on concatenated features, producing a Gaussian-enhanced 3D representation for improved detection.}
\label{fig_framework}
\end{figure*}
\textbf{Problem definition. }Given a set of RGB images $\{\mathbf{I}_i \in \mathbb{R}^{H \times W \times 3}\}_{i=1}^{n}$ captured from different viewpoints with known camera intrinsic parameters $\{\mathbf{K}_i\in \mathbb{R}^{3 \times 4}\}_{i=1}^n$ and camera extrinsic parameters $\{\mathbf{P}_i\in \mathbb{R}^{4 \times 4}\}_{i=1}^n$, multi-view 3D object detection aims to predict 3D bounding boxes $\mathcal{B} = {\{\mathbf{c}_j, \mathbf{b}_j\}}_{j=1}^M$ for all objects in the scene, where $\mathbf{c}_j$ denotes the object category and $\mathbf{b}_j$ represents the 3D box parameters including center location, dimensions, and orientation. 

\textbf{Overall pipeline.} Our GVSynergy-Det framework, illustrated in Figure \ref{fig_framework}, consists of four main components. First, the 2D-to-3D Feature Lifting module constructs a 3D voxel representation $\mathbf{V}$ by back-projecting multi-view features into 3D space. Second, the Generalizable 3D Gaussian Splatting module extracts depth and produces Gaussian representations containing rich geometric and semantic information. Third, the Cross-Representation Enhancement module synergistically combines the Gaussian features with the standard 3D voxel features through an adaptive mechanism. Finally, the Detection Head performs object localization and classification on the enhanced features. The key innovation lies in how we leverage Gaussian Splatting not just for depth regularization, but as a complementary representation that enriches the entire detection pipeline through cross-representation integration.

\subsection{2D-to-3D Feature lifting} \label{2dto3d}
This module transforms 2D image features into a 3D voxel representation $\mathbf{V}$ through a two-stage process: 2D image feature extraction, followed by geometric back-projection into 3D space.

\textbf{2D image feature extraction.} We employ a small (21M) Transformer backbone \cite{oquab2023dinov2} to extract 2D features from input images. Given an input image $\mathbf{I}_i$, we obtain patch tokens $\mathbf{h}_i \in \mathbb{R}^{N \times D}$ and class tokens $\mathbf{c}_i \in \mathbb{R}^{1 \times D}$. We concatenate the class tokens with the patch tokens to incorporate global context and generate the image features as follows:
\begin{equation}
    \mathbf{f}_i = \text{Proj}([\text{Repeat}(\mathbf{c}_i, N), \mathbf{h}_i]),
\end{equation}
where $\text{Repeat}(\mathbf{c}_i, N)$ expands the class token to $\mathbb{R}^{N \times D}$ by repeating it $N$ times, and Proj is a linear projection followed by activation. 
\textbf{3D feature volume construction. }Given the multi-view 2D features $\{\mathbf{f}_i\}_{i=1}^n$, we construct a 3D voxel representation through back-projection.
First, we establish a 3D coordinate system with the vertical axis perpendicular to the ground plane and two orthogonal horizontal axes.  

Within this coordinate system, we define a voxelized representation consisting of $N_x \times N_y \times N_z$ discrete cells. Each voxel center $\mathbf{p} = (x, y, z)^T$ is transformed into image coordinates through the calibrated projection matrices:
\begin{equation}
    \tilde{\mathbf{u}}_i = \mathbf{S} \cdot \mathbf{K}_i \cdot \mathbf{P}_i \cdot [\mathbf{p}^T, 1]^T, 
\end{equation}
where $\tilde{\mathbf{u}}_i = [u_i, v_i, 1]^T$ represents homogeneous pixel coordinates in view $i$, $\mathbf{K}_i$ and $\mathbf{P}_i$ denote the intrinsic and extrinsic camera parameters respectively, and $\mathbf{S}$ is a scaling matrix that accounts for the resolution difference between the feature maps and original images.

The feature vector associated with each voxel is obtained by sampling from the 2D feature maps at the projected locations:
\begin{equation}
    \mathbf{f}_{\mathbf{p}}^i = \text{sample}(\mathbf{f_i}, (u_i, v_i)),
\end{equation}
where the sampling operation uses bilinear interpolation to handle non-integer coordinates.

To handle visibility constraints and field-of-view limitations, we compute binary validity indicators $m_{\mathbf{p}}^i$ that equal 1 when the projected coordinates fall within the image boundaries and 0 otherwise. The final voxel features are computed as a weighted average across all contributing views:
\begin{equation}
    \mathbf{V}_{\mathbf{p}} = \frac{\sum_{i=1}^{n}m_{\mathbf{p}}^i \cdot \mathbf{f}_{\mathbf{p}}^i}{\sum_{i=1}^n m_{\mathbf{p}}^i}.
\end{equation}
Voxels receiving no valid projections are assigned zero features. This process yields a dense 3D feature volume $\mathbf{V} \in \mathbb{R}^{C_f \times N_x \times N_y \times N_z}$ that encodes multi-view information in a unified 3D representation, ready for subsequent processing by the detection pipeline.

\subsection{Pixel-aligned 3D Gaussian Splatting}
To enable efficient novel view synthesis while maintaining consistency with the 3D detection task, we incorporate a pixel-aligned 3D Gaussian Splatting module that generates view-consistent 3D representations directly from multi-view features.

\textbf{Depth estimation.}  The image features $\mathbf{f}_i$ are first processed through a reassembly module that restores spatial dimensions and adjusts feature resolutions to generate the depth features $\mathbf{F}_i$. Then we employ a depth estimation head consisting of convolutional layers that operate on the 2D features $\mathbf{F_i}$ to predict depth maps. For each view $i$, we establish 3D positions for all Gaussian centers by unprojecting pixel coordinates using the predicted depth.

\textbf{Gaussian parameter prediction.} Beyond spatial positions, we predict the remaining per-pixel Gaussian attributes from the 2D depth features $\mathbf{F}_i$ using a lightweight Gaussian regressor. The Gaussian regressor consists of convolutional layers that output 66 channels for each pixel. Specifically, we decompose these 66 channels into three distinct components: 1 channel for opacity $\alpha \in [0,1]$ (activated by sigmoid), 1 channel for the fusion weight $w \in [0,1]$ (activated by sigmoid), and 64 channels for the high-dimensional latent feature vector $\mathbf{h}_g \in \mathbb{R}^{64}$. The latent feature $\mathbf{h}_g$ encodes high-level geometric and appearance information, which is subsequently decoded into spherical harmonics, rotation, and scaling parameters.

\textbf{Multi-view Gaussian fusion.} We employ an iterative multi-view fusion strategy adapted from \cite{10.5555/3737916.3741325} to merge redundant Gaussians and construct a unified scene representation. Starting with Gaussians from the first view, we incrementally integrate subsequent views. For each new view, we project the current global Gaussians into its coordinate system and check for geometric consistency based on depth differences. If a new Gaussian aligns with an existing global Gaussian, we fuse their latent features $\mathbf{h}_g$ using a Gated Recurrent Unit (GRU) \cite{Cho2014LearningPR} guided by their fusion weights $w$. Conversely, if a Gaussian represents a newly observed region, it is appended to the global set. This mechanism effectively eliminates redundancy while preserving unique geometric details from all viewpoints.

The resulting Gaussian representation serves dual purposes in our framework. First, it provides additional photometric supervision through differentiable rendering, helping to refine the depth estimates and features. Besides, it offers an alternative 3D representation that can be queried alongside our voxel features for enhanced detection performance.

\subsection{Cross-Representation Enhancement}

While both voxel-based and Gaussian-based representations offer unique advantages for 3D understanding, they capture complementary information. Voxels provide regular grid structures ideal for convolution-based detection, while Gaussians offer continuous, view-consistent representations with fine-grained geometric details. To leverage both modalities synergistically, we propose a cross-representation enhancement module that enriches voxel features with complementary information from the Gaussian representation.

\textbf{Gaussian voxelization. }To enable integration between the two representations, we first transform the irregular Gaussian primitives into a regular voxel grid aligned with our detection backbone. Given the Gaussian centers and their associated features $\mathbf{H}_g \in \mathbb{R}^{N_g \times C_g}$ (where $C_g=64$), we perform spatial binning:
\begin{equation}
    \mathbf{v}_{x,y,z} = \floor*{\frac{\mathbf{m}-\mathbf{o}}{s_v}}, 
\end{equation}
where $\mathbf{m}$ is a Gaussian's 3D position, $\mathbf{o}$ is the scene origin, and $s_v$ is the voxel size. For efficient aggregation, we linearize the 3D indices and accumulate features within each voxel via average pooling:
\begin{equation}
    \mathbf{V}_g[\mathbf{v}]=\frac{1}{|\mathcal{G}_\mathbf{v}|}\sum_{j \in \mathcal{G}_\mathbf{v}}{\mathbf{h}^j_g},
\end{equation}
where $\mathcal{G}_\mathbf{v}$ represents the set of Gaussians falling into voxel $\mathbf{v}$. Crucially, we also compute an occupancy mask $\mathbf{O} \in \{0,1\}^{N_x \times N_y \times N_z}$ indicating which voxels contain at least one Gaussian primitive, providing explicit geometric guidance for the integration process.

Then, we encode the Gaussian features as follows:
\begin{equation}
    \hat{\mathbf{V}}_g = \mathcal{P}_g(\mathbf{V}_g),
\end{equation}
where $\mathcal{P}_g$ is a Gaussian Feature Encoder that consists of stacked convolutional layers that transform the 64-dimensional Gaussian features to 256 dimensions while preserving spatial structure.

The encoded features are then modulated by the occupancy mask to explicitly indicate which voxels contain geometric information. This masking operation ensures that empty voxels do not contribute noise to the integration process and helps the network distinguish between genuinely empty space and regions with features.
\begin{equation}
    \hat{\mathbf{V}}_g^{o} = \hat{\mathbf{V}}_g \otimes \mathbf{O},
\end{equation}

\textbf{Adaptive cross-enhancement. }Given the voxelized encoded Gaussian features $\hat{\mathbf{V}}_g^{o}$ and the original voxel features $\mathbf{V}$ from our detection branch, we design an adaptive enhancement mechanism that learns to integrate both representations based on their local reliability and occupancy information.

The integration weights are computed adaptively based on the concatenated features:
\begin{equation}
    [\mathbf{\alpha}_v, \mathbf{\alpha}_g] = \text{softmax}(\mathcal{W}([\mathbf{V}, \hat{\mathbf{V}}_g^{o}])),
\end{equation}
where $\mathcal{W}$ is a lightweight weight network that outputs two weight channels. These weights allow the model to dynamically balance between voxel and Gaussian information based on factors such as feature quality, spatial coverage, and occupancy patterns.

The final enhanced features are computed through adaptive weighting followed by refinement:
\begin{equation}
    \mathbf{V}_{\text{e}} = \mathcal{F}([\mathbf{\alpha}_v \otimes \mathbf{V}, \mathbf{\alpha}_g \otimes \hat{\mathbf{V}}_g^{o}]), 
\end{equation}
where $\mathcal{F}$ is a convolutional fusion network.

This cross-representation enhancement enriches the voxel features with fine-grained geometric details from the Gaussian representation, particularly benefiting regions with sparse voxel coverage or complex geometric structures. The occupancy-aware design ensures that the enhancement process is guided by actual geometric presence, resulting in more robust features for downstream 3D detection.

\subsection{Detection Head}
Following the design principles of ImVoxelNet \cite{rukhovich2022imvoxelnet}, we employ a 3D feature pyramid network (FPN) as the neck to process the enhanced voxel features, followed by a multi-branch 3D detection head. The 3D FPN extracts multi-scale features from the voxel representation through 3D convolutions and upsampling operations. The detection head consists of three parallel branches operating on the multi-scale voxel features: a center branch that predicts a centerness score for each voxel location, a regression branch that estimates 3D bounding box parameters, including distances to box faces and orientation angle, and a classification branch that predicts object category probabilities for each voxel. 

Our training objective combines detection losses with a novel rendering supervision from the Gaussian representation:
\begin{equation}
    \mathcal{L}_{\text{total}} = \mathcal{L}_{\text{det}} + \lambda_{\text{render}} \mathcal{L}_{\text{render}},
\end{equation}
where $\lambda_{\text{render}}$ is a weighting factor balancing the two objectives.

Following ImVoxelNet, the detection loss consists of three components corresponding to the three prediction branches:
\begin{equation}
    \mathcal{L}_{\text{det}} = \mathcal{L}_{\text{center}} +  \mathcal{L}_{\text{bbox}} + \mathcal{L}_{\text{cls}},
\end{equation}
where $\mathcal{L}_{\text{center}}$ uses binary cross-entropy with sigmoid activation to supervise the centerness predictions, $\mathcal{L}_{\text{cls}}$ employs focal loss to address the severe foreground-background imbalance in 3D voxel grids, and $\mathcal{L}_{\text{bbox}}$ applies a rotated IoU loss between predicted and ground truth boxes. 

To leverage the view synthesis capability of our Gaussian representation, we introduce an additional photometric supervision. The render loss $\mathcal{L}_{\text{render}}$ is computed by the MSE loss between the rendered images from the predicted Gaussians and the ground truth target images.

\section{Experiments}
\subsection{Experimental Setting}
\textbf{Datasets.} We evaluate our method on two challenging indoor 3D object detection benchmarks. ScanNetV2 \cite{dai2017scannet} is a large-scale RGB-D dataset containing 1,513 indoor scenes with dense semantic and instance annotations for 18 object categories. The dataset provides multi-view RGB images, depth maps, and camera parameters, making it ideal for evaluating multi-view 3D detection methods. ARKitScenes \cite{dehghan2021arkitscenes} is a more recent dataset captured using Apple's ARKit framework, consisting of 5,047 indoor scenes with RGB images and camera trajectories. It includes annotations for 17 object categories and presents additional challenges due to its diverse capture conditions and larger scene variations compared to ScanNetV2.


\begin{table*}[t]
    \centering
\caption{\textbf{Comparison with state-of-the-art methods on ScanNetV2.} Our method achieves the best performance among approaches without dense 3D supervision.}
    \label{tab:model_comparison}
    \begin{tabular}{l|c|c|c|c|c}
    \toprule
    \multirow{2}{*}{\textbf{Model}} & \multirow{2}{*}{\textbf{Train/Test Views}} & \multirow{2}{*}{\textbf{Venue}}  & \multirow{2}{*}{\textbf{3D Supervision}} & \multirow{2}{*}{\textbf{mAP@0.25}} & \multirow{2}{*}{\textbf{mAP@0.50}} \\
    & & & &  & \\ 
    \midrule
    ImGeoNet \cite{tu2023imgeonet}  & 20/50 & ICCV 2023  & Point clouds & 54.5 & 28.9 \\
    CN-RMA \cite{cnmar2024} & 20/50 & CVPR 2024  & TSDF volume & 51.0 & 31.0 \\
    3DGeoDet \cite{11045444} & 20/50 & TMM 2025 & Depth maps & 59.6 & 34.3 \\
    \midrule 
    ImVoxelNet \cite{rukhovich2022imvoxelnet} & 20/50 & WACV 2022  & \XSolidBrush & 43.4 & 19.9 \\
    NeRF-Det \cite{Xu_2023_ICCV} & 20/50 & ICCV 2023  & \XSolidBrush & 50.4 & 25.2 \\
    MVSDet \cite{10.5555/3737916.3742138} & 20/50 & NeurIPS 2024  & \XSolidBrush & 54.0 & 29.0 \\
    NeRF-Det++ \cite{10969554} & 50/50 & TIP 2025  & \XSolidBrush & 53.9 & 29.6 \\
    \textbf{Ours} & 20/50 & -  & \XSolidBrush & \textbf{56.3} & \textbf{32.1} \\
    \bottomrule

    \end{tabular}
\end{table*}

\begin{table*}[t] \centering
    \caption{\textbf{3D detection performance with different testing views on the ScanNetV2 dataset.} 20 views are used for training for all approaches.}
    \label{tab:scannetv2_views}
    \begin{tabular}{l|c|c|c|c|c|c|c|c}
    \toprule
    \multirow{2}{*}{Method} &
    \multicolumn{8}{c}{Performance (mAP@0.25 / mAP@0.50) $\uparrow$} \\
    \cmidrule{2-9}
       & 5 views & 10 views & 20 views & 30 views & 40 views & 50 views & 60 views & 70 views\\
    \midrule
    ImVoxelNet \cite{rukhovich2022imvoxelnet}  & 25.6 / 9.70 & 34.0 / 13.3 & 40.7 / 17.4 & 40.8 / 18.7 & 44.3 / 20.7 & 43.4 / 19.9 & 41.6 / 19.1 & 43.8 / 21.5\\
    NeRF-Det \cite{Xu_2023_ICCV}  & 24.2 / 9.30 & 38.7 / 15.9 & 44.8 / 21.7 & 49.1 / 25.4 & 50.4 / 24.9 & 50.4 / 25.2 & 52.3 / 26.0 & 52.6 / 26.8 \\
    MVSDet \cite{10.5555/3737916.3742138}  & 31.5 / 12.7  & 42.1 / 19.6 & 49.6 / 25.0 & 52.1 / 27.6 & 53.1 / 27.9 & 54.0 / 29.0 & 54.1 / 28.6 & 54.6 / 29.4 \\
     & \textbf{32.1} / \textbf{15.0} & \textbf{45.7} / \textbf{23.7} & \textbf{52.8} / \textbf{28.8} & \textbf{54.6} / \textbf{31.2} & \textbf{56.1} / \textbf{31.5} & \textbf{56.3} / \textbf{32.1} & \textbf{57.0} / \textbf{31.2} & \textbf{56.1} / \textbf{31.7}\\
    \multirow{-2}{*}{\textbf{Ours}}
     & \textbf{+0.6} / \textbf{+2.3} & \textbf{+3.6} / \textbf{+4.1} & \textbf{+3.2} / \textbf{+3.8} & \textbf{+2.5} / \textbf{+3.6} & \textbf{+3.0} / \textbf{+3.6} & \textbf{+2.3} / \textbf{+3.1} & \textbf{+2.9} / \textbf{+2.6} & \textbf{+1.5} / \textbf{+2.3}\\
    \bottomrule
    \end{tabular}
\end{table*}

\textbf{Evaluation metric and compared method. }Following standard protocols in 3D object detection, we report mean Average Precision (mAP) at IoU thresholds of 0.25 and 0.5. We compare our method against state-of-the-art multi-view 3D detection approaches, including methods with expensive dense 3D supervision (ImGeoNet \cite{tu2023imgeonet}, CN-RMA \cite{cnmar2024}, 3DGeoDet \cite{11045444}) and those without explicit 3D supervision (ImVoxelNet \cite{rukhovich2022imvoxelnet}, NeRF-Det \cite{Xu_2023_ICCV}, MVSDet \cite{10.5555/3737916.3742138}, NeRF-Det++ \cite{10969554}). 

\textbf{Implementation details.} Our implementation is based on the MMDetection3D framework. The voxel representation has a shape of 40×40×16 with a voxel size of 0.16m, covering a spatial volume suitable for indoor scenes. For ScanNetV2, we train for 9 epochs with a batch size of 1, using the AdamW optimizer with an initial learning rate of 1e-4, which is decayed by a factor of 0.1 at the 8th epoch. For ARKitScenes, we extend training to 10 epochs with the same learning rate and decay schedule, accounting for the dataset's larger scale and complexity. All experiments are conducted on 4 NVIDIA RTX 6000 Ada GPUs. 

\subsection{Comparison with State-of-the-art Methods}

\textbf{ScanNetV2. }As shown in Table \ref{tab:model_comparison}, our GVSynergy-Det achieves competitive performance on the ScanNetV2 benchmark despite using no dense geometry supervision. Among methods relying only on 3D bounding boxes supervision, we achieve the best performance with 56.3 mAP@0.25 and 32.1 mAP@0.5, outperforming the recent MVSDet by 2.3 and 3.1, respectively. Notably, our method even surpasses some approaches using dense 3D geometry supervision, such as ImGeoNet, demonstrating the effectiveness of our cross-representation detection model.

Table \ref{tab:scannetv2_views} presents a comprehensive evaluation of 3D detection performance across varying numbers of test views on ScanNetV2. Our method consistently outperforms all baselines across different view counts, demonstrating superior robustness and scalability. With limited views, our approach already shows notable improvements, achieving 32.1 mAP@0.25 with just 5 views. As the number of views increases, the performance gap becomes more pronounced. At 50 test views, we achieve 56.3 mAP@0.25 and 32.1 mAP@0.50, surpassing the second-best method MVSDet by 2.3 and 3.1, respectively. The consistent improvements across all view settings, particularly the substantial gains of up to 3.0 mAP@0.25 and 3.6 mAP@0.50 at 40 views, validate the effectiveness of our model in leveraging multi-view information more efficiently than existing approaches.

\begin{figure}[t]
\centering
\includegraphics[width=0.95\linewidth]{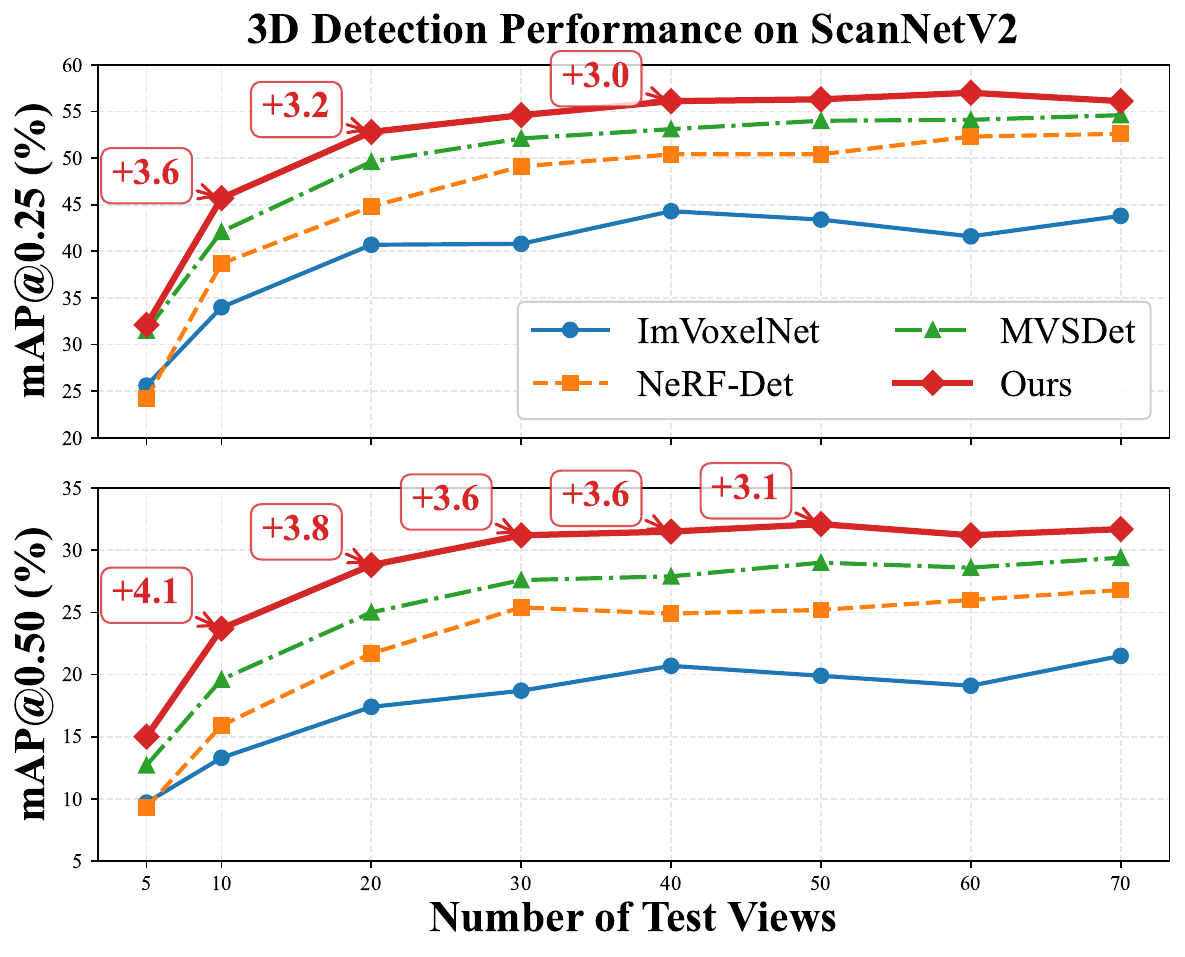} 
\caption{\textbf{3D Detection Performance on ScanNetV2 with varying test views.} We plot mAP@0.25 (top) and mAP@0.50 (bottom) against the number of input views ranging from 5 to 70. GVSynergy-Det (red line) consistently outperforms competing methods, demonstrating superior robustness particularly in sparse-view scenarios.}
\label{fig:views_curve}
\end{figure}

To provide a more intuitive understanding of model robustness, we visualize the performance trends across varying view counts in Figure \ref{fig:views_curve}. While Table \ref{tab:scannetv2_views} lists the specific numerical results, the curves highlight the rate of performance degradation as views decrease. It is evident that GVSynergy-Det (red curve) maintains a significant margin over MVSDet (green dashed line) and NeRF-Det (orange line), particularly in the challenging sparse-view regime (10 to 40 views). As the difficulty increases, except for the case of 5 views, the gap between our method and the baselines widens, which validates that our cross-representation enhancement can effectively generate missing geometry when visual overlap is limited.

\begin{table}[t]
    \centering
    \caption{\textbf{Comparison of various methods on ARKitScenes.} We randomly choose 50 views for training, and 100 views for testing.}
    \label{tab:arkitscenes_comparison}
    \begin{tabular}{l|c|c|c}
    \toprule
    \textbf{Model}  & \textbf{3D Sup.} & \textbf{mAP@0.25} & \textbf{mAP@0.50} \\
    \midrule
    ImGeoNet \cite{tu2023imgeonet}   & Point clouds & 60.2 & 43.4 \\
    CN-RMA \cite{cnmar2024}  & TSDF volume & 67.6 & 56.5 \\
    \midrule 
    ImVoxelNet \cite{rukhovich2022imvoxelnet}   & \XSolidBrush & 27.3 & 4.3 \\
    NeRF-Det \cite{Xu_2023_ICCV}   & \XSolidBrush & 39.5 & 21.9 \\
    MVSDet \cite{10.5555/3737916.3742138}  & \XSolidBrush & 42.9 & 27.0 \\

    \textbf{Ours} & \XSolidBrush & \textbf{44.1} & \textbf{30.6} \\
    \bottomrule
    \end{tabular}
\end{table}

\textbf{ARKitScenes. }Table \ref{tab:arkitscenes_comparison} presents our results on the ARKitScenes benchmark, which poses additional challenges with its diverse real-world indoor scenes captured using mobile devices. Despite these challenges, our method achieves 44.1 mAP@0.25 and 30.6 mAP@0.50, establishing new state-of-the-art results among approaches without dense geometry supervision. We outperform MVSDet by 1.2 mAP@0.25 and 3.6 mAP@0.50, demonstrating significant improvements in precise localization.

\subsection{Efficiency and Resource Analysis}
To demonstrate the practical deployability of GVSynergy-Det, we conduct a comprehensive analysis of computational efficiency, memory consumption, and model complexity. Table \ref{tab:efficiency} presents the inference speed (measured in Scenes/sec with 50 views per scene), GPU memory usage, parameter count, and detection performance on the ScanNetV2 dataset.

\begin{table*}[t]
\centering
\caption{Comparison of efficiency, resource consumption, and detection performance on ScanNetV2. All methods are evaluated with 50 input views per scene. Our method achieves the best trade-off, significantly outperforming MVSDet in memory efficiency and accuracy while maintaining a compact model size.}
\label{tab:efficiency}
\setlength{\tabcolsep}{12pt} 
\begin{tabular}{lccccc}
\toprule
\textbf{Method} & \textbf{Speed} & \textbf{Memory} & \textbf{Model Size} & \textbf{mAP@0.50} & \textbf{mAP@0.25} \\
& (Scenes/sec) & (GB) & (M) & (\%) & (\%) \\
\midrule
NeRF-Det \cite{Xu_2023_ICCV} & \textbf{3.6} & \textbf{7} & 128 & 25.2 & 50.4 \\
MVSDet \cite{10.5555/3737916.3742138} & 1.8 & 16 & 109 & 29.0 & 54.0 \\
\textbf{Ours} & 2.2 & 8 & \textbf{107} & \textbf{32.1} & \textbf{56.3} \\
\bottomrule
\end{tabular}
\end{table*}

As detailed in Table \ref{tab:efficiency}, existing methods struggle to balance performance with resource usage. NeRF-Det offers high inference speed (3.6 Scenes/sec) and low memory usage (7 GB) but fails to capture precise geometry, resulting in lower detection accuracy (25.2\% mAP@0.50). On the other hand, MVSDet achieves respectable accuracy but incurs a heavy computational burden, requiring 16 GB of GPU memory—double that of our method—and suffering from the slowest inference speed (1.8 Scenes/sec) due to its expensive plane-sweep volume construction.

In contrast, GVSynergy-Det achieves an optimal balance. By synergizing voxel and Gaussian representations, we attain state-of-the-art accuracy, surpassing MVSDet by 3.1 in mAP@0.50 and 2.3 in mAP@0.25 while maintaining a compact model size (107 M) and a low memory footprint (8 GB). Figure \ref{fig:model_size_vs_perf} visualizes this advantage, plotting model size against mAP@0.25. Our approach occupies the ideal top-left region of the performance frontier, demonstrating that our adaptive cross-enhancement module effectively extracts rich geometric details without the need for the excessive memory overhead seen in previous Gaussian-based or NeRF-based approaches.

\subsection{Ablation Study}
\begin{table*}[t]
    \centering
    \caption{\textbf{Ablation study on ScanNetV2.} We start with a Voxel-only baseline and incrementally add our proposed components. The results show that while auxiliary Gaussian supervision is helpful, the most significant gains come from our adaptive feature integration strategy.}
    \label{tab:ablation_scannet}
    \renewcommand{\theadalign}{cc} 
    \renewcommand{\theadfont}{\bfseries} 
    \begin{tabular}{l|ccc|cc|cc}
    \toprule
     \thead{Configuration} & \thead{Gaussian \\ Supervision} & \thead{Feature \\ Fusion} & \thead{Adaptive \\ Mechanism}  & \thead{PSNR} & \thead{SSIM} & \thead{mAP@0.25} & \thead{mAP@0.50} \\
    \midrule
    Baseline & \XSolidBrush & \XSolidBrush & \XSolidBrush & - & - & 53.7 & 28.6 \\

     + Auxiliary Loss & \Checkmark & \XSolidBrush & \XSolidBrush & \textbf{16.2} & \textbf{0.57} & 53.9 & 29.4 \\
     
     + Direct Fusion & \Checkmark & \Checkmark & \XSolidBrush & 16.1 & 0.56 &55.1 & 31.0 \\ 

     Full Model & \Checkmark & \Checkmark & \Checkmark & 16.1  & 0.56 &\textbf{56.3} & \textbf{32.1} \\
    \bottomrule
    \end{tabular}
\end{table*}

\begin{table}[t]
    \centering
    \caption{\textbf{Ablation study on the ARKitScenes dataset.} The trend confirms that our cross-representation enhancement module consistently improves detection performance across different datasets.}
    \label{tab:ablation_arkit}
    \renewcommand{\theadalign}{cc} 
    \renewcommand{\theadfont}{\bfseries} 
    \begin{tabular}{cc|cc}
    \toprule
     \thead{Gaussian \\ Supervision} & \thead{Cross Rep \\ Enh.} & \thead{mAP@0.25} & \thead{mAP@0.50} \\
    \midrule
     \XSolidBrush & \XSolidBrush  & 41.7 & 27.0 \\

      \Checkmark & \XSolidBrush  & 43.7 & 28.3 \\

      \Checkmark & \Checkmark  &\textbf{44.1} & \textbf{30.6} \\
    \bottomrule
    \end{tabular}
\end{table}

To validate the effectiveness of our core contributions, we conduct detailed ablation studies on both the ScanNetV2 and ARKitScenes datasets.

\textbf{Ablation on ScanNetV2.} The primary ablation, presented in Table \ref{tab:ablation_scannet}, was performed on ScanNetV2 using 20 training and 50 test views. Our analysis begins with a voxel-only baseline, which achieves 53.7 mAP@0.25. Introducing the Gaussian branch for auxiliary supervision via $\mathcal{L}_{\text{render}}$ provides a boost on mAP@0.25 and mAP@0.50, demonstrating the benefit of geometric regularization. A more substantial improvement is gained by directly using our Fusion Network to fuse the features from both branches, which proves that explicitly providing the detector with 3D Gaussian features is highly effective. The final and most critical step is activating the adaptive mechanism in our full model, which provides a further performance leap to 56.3 mAP@0.25 and 32.1 mAP@0.50. This confirms that the learnable, adaptive integration is the key component enabling synergy between the two representations. We also note that as detection performance improves, rendering quality (PSNR/SSIM) remains stable, suggesting the model learns to prioritize features for detection, a reasonable trade-off for the primary task.

\textbf{Ablation on ARKitScenes.} To confirm that these architectural benefits generalize, we performed a second ablation study on the ARKitScenes dataset, with the results shown in Table \ref{tab:ablation_arkit}. The results exhibit a trend consistent with our findings on ScanNetV2. Starting from a baseline of 41.7 mAP@0.25, adding auxiliary supervision provides a clear benefit (+2.0 mAP@0.25). The full model, incorporating our complete fusion module, achieves the best performance at 44.1 mAP@0.25 and 30.6 mAP@0.50. This consistent improvement across a different dataset underscores the robustness and effectiveness of our proposed cross-representation enhancement module.

\subsection{Qualitative Analysis}

We present a comprehensive visual analysis of our method's performance. We first compare GVSynergy-Det against state-of-the-art baselines on ScanNetV2, followed by an analysis of detection fidelity on the ARKitScenes dataset.

\subsubsection{Comparison on ScanNetV2}
\begin{figure*}[t]
\centering
\includegraphics[width=1.0\linewidth]{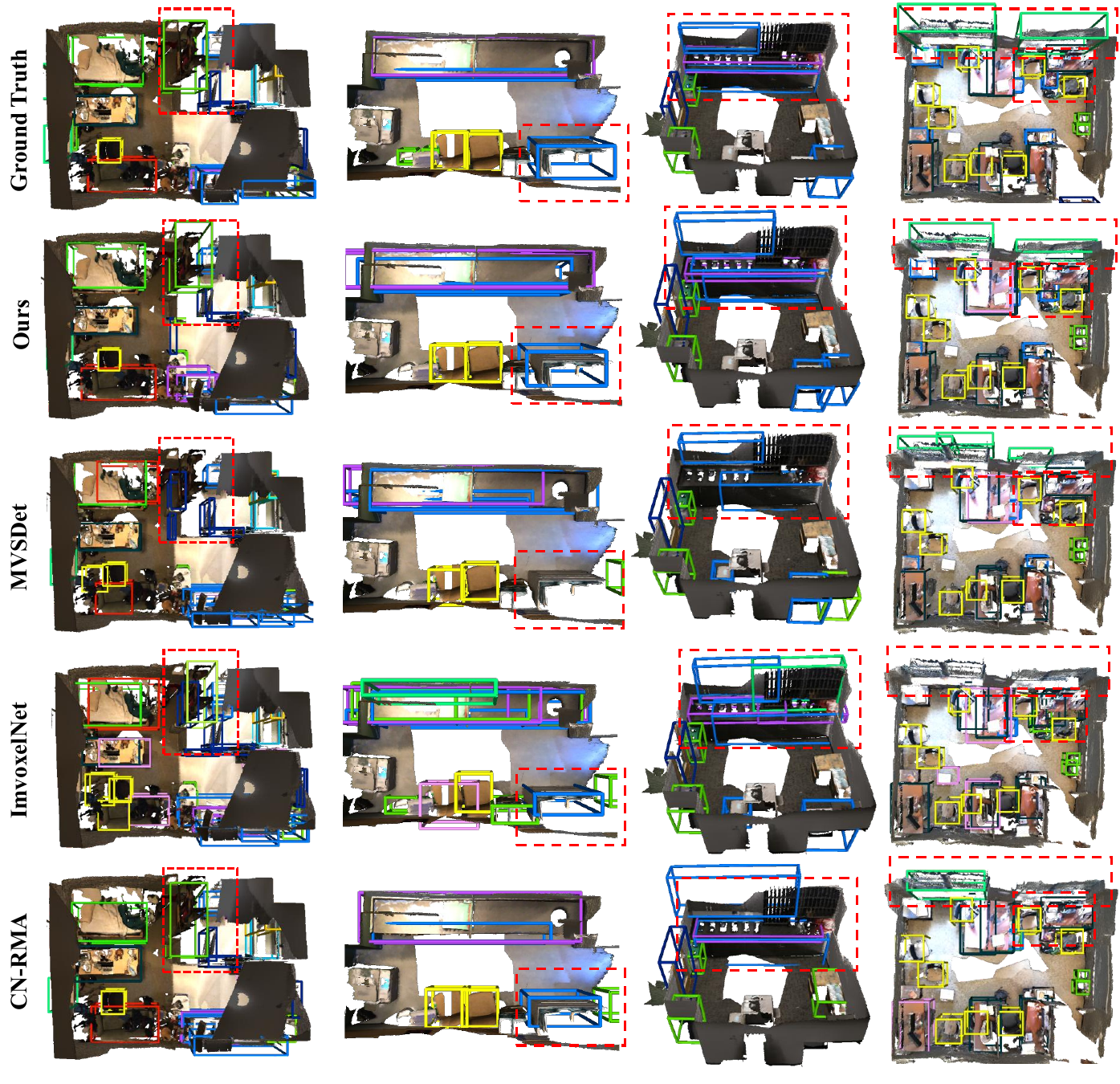}
\caption{\textbf{Qualitative comparison of 3D object detection on ScanNetV2.} The rows correspond to (from top to bottom): Ground Truth, Ours, MVSDet, ImVoxelNet, and CN-RMA. The dashed red boxes highlight specific challenging regions. In the second column, baselines hallucinate an extra object next to the cabinet (blue box), whereas our method correctly predicts only the single cabinet. In the fourth column, our method correctly identifies exactly two windows, whereas baselines predict incorrect counts or misaligned boxes.}
\label{fig:qualitative_scannet}
\end{figure*}

Figure \ref{fig:qualitative_scannet} illustrates the comparison between GVSynergy-Det, the Ground Truth, and three competitive baselines (MVSDet, ImVoxelNet, and CN-RMA) on the ScanNetV2 validation set.

\textbf{Suppression of false positives:}
The second column of Figure \ref{fig:qualitative_scannet} demonstrates the model's precision. As shown in the red dashed region, the Ground Truth contains a single object: a cabinet (marked by the blue bounding box). While the baseline methods detect the cabinet, they all suffer from over-detection, hallucinating an additional, non-existent object adjacent to the target. In contrast, our method exhibits superior noise suppression. It correctly identifies the single cabinet without generating any false positive bounding boxes, perfectly matching the Ground Truth.

\textbf{Geometric precision:}
The fourth column highlights the ability to resolve structural elements. The target area contains exactly two windows. Existing methods fail to capture the correct topology, predicting incorrect counts (0, 1, or 3 instances) or imprecise boxes. GVSynergy-Det correctly identifies exactly two windows and localizes them with high precision.

\subsubsection{Performance on ARKitScenes}
\begin{figure}[t]
\centering
\includegraphics[width=1.0\linewidth]{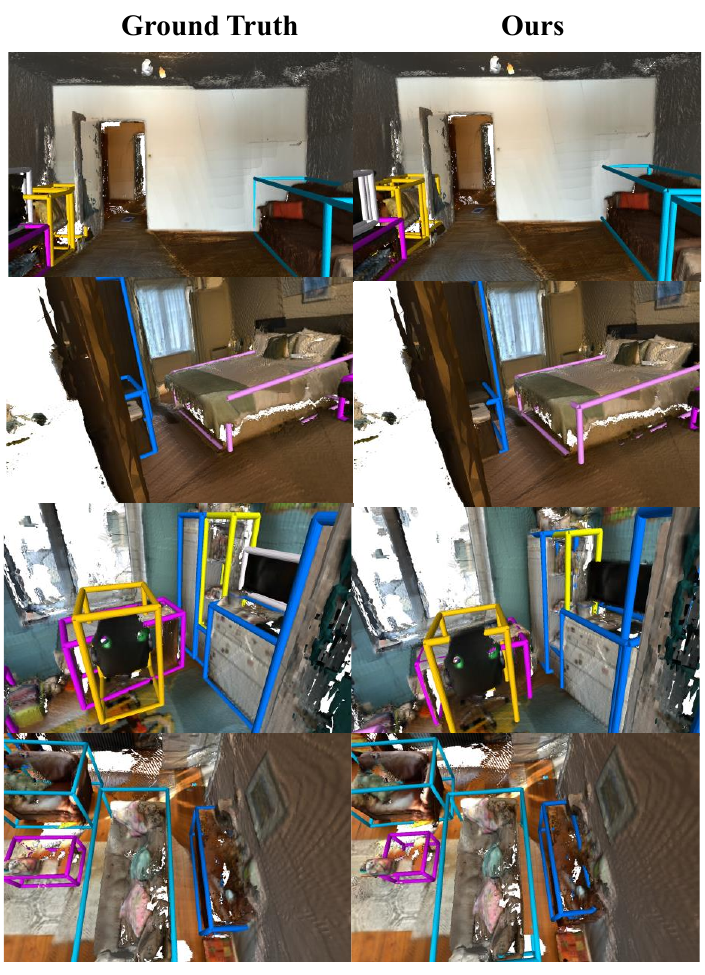}
\caption{\textbf{Qualitative results on the ARKitScenes dataset.} The left column shows the Ground Truth annotations, and the right column shows the predictions from our method. Our model demonstrates high detection fidelity, producing bounding boxes that accurately match the scale, orientation, and location of the Ground Truth objects.}
\label{fig:qualitative_arkit}
\end{figure}

To validate the versatility of our approach across different data distributions, we further evaluate our method on the ARKitScenes dataset. Figure \ref{fig:qualitative_arkit} presents the qualitative results.

As observed in Figure \ref{fig:qualitative_arkit}, GVSynergy-Det achieves high detection fidelity on this benchmark. Despite the differences in sensor characteristics and scene layouts compared to ScanNetV2, our method effectively learns the geometric representations required for accurate 3D detection. The predicted bounding boxes exhibit tight alignment with the Ground Truth, successfully recovering object orientations and dimensions in complex indoor scenarios.

\section{Conclusion}
In this paper, we presented GVSynergy-Det, a novel framework for multi-view 3D object detection that synergistically combines continuous Gaussian and discrete voxel representations. Our key contribution is demonstrating that these complementary geometric representations can be effectively integrated at the feature level through adaptive cross-enhancement, rather than treating them as separate modules. By developing a learnable integration mechanism that dynamically balances contributions from both representations based on local geometric occupancy and feature reliability, we achieve significant improvements over existing methods without requiring any depth or point cloud supervision.

\bibliographystyle{IEEEtran} 
\bibliography{ref}

\end{document}